\documentclass{article} % For LaTeX2e
\usepackage{iclr2027_conference,times}

\usepackage{amsmath,amsfonts,bm}
\usepackage{hyperref}
\usepackage{url}
\usepackage{graphicx}
\usepackage{enumitem}
\usepackage{booktabs}
\usepackage{multirow}
\usepackage{fontawesome5}

\def\eqref#1{equation~\ref{#1}}
\def\1{\bm{1}}

\DeclareMathAlphabet{\mathsfit}{\encodingdefault}{\sfdefault}{m}{sl}
\SetMathAlphabet{\mathsfit}{bold}{\encodingdefault}{\sfdefault}{bx}{n}

\title{Pretrain Once, Route Anywhere: Towards a Foundation Model for LLM Routing}

\author{%
Guannan Lai \qquad Han-Jia Ye \\
School of Artificial Intelligence, Nanjing University\\
National Key Laboratory for Novel Software Technology, Nanjing University \\
\texttt{\{laign, yehj\}@lamda.nju.edu.cn}
}

\iclrfinalcopy

\begin{document}

\maketitle
\lhead{Preprint}

\begin{abstract}
Large language model (LLM) routing aims to assign each query to the most
suitable model from a heterogeneous candidate pool, improving the
quality--efficiency trade-off of LLM inference. Existing routers are typically
learned through \emph{local fitting}: a router is optimized for a particular
query workload and candidate pool, and often requires additional supervision
or retraining as the routing environment changes. We ask whether LLM routing
can instead be approached from a \textbf{foundation-model} perspective,
learning a reusable routing capability that generalizes across tasks,
candidate models, and deployment conditions. To this end, we introduce
\textbf{RouteFM}, which learns to characterize anonymous candidate models from
behavioral context and infer their target-specific capabilities, rather than
binding routing decisions to fixed model identities or a single environment.
Through episodic pretraining across heterogeneous routing environments, this
capability can be reused by a frozen router and adapted to new environments
through context alone. Experiments demonstrate transfer across changes in
domains, modalities, candidate pools, and context budgets, with the largest
gains when behavioral evidence is limited. On MMR-Bench, which is excluded
from pretraining, RouteFM outperforms the strongest baseline by 2.23 quality
points with only eight observations per candidate. These results support
moving LLM routing from repeated local fitting toward a
\emph{pretrain once, route anywhere} paradigm.
Our code is publicly available at
\url{https://github.com/LAMDA-Model-Reuse/RouteFM}.
\end{abstract}

\section{Introduction}

Large language model (LLM) routing
\citep{chen2024frugalgpt,ong2024routellm} dynamically assigns each query to
a suitable model from a candidate pool with heterogeneous capabilities,
inference costs, and latency. This provides a practical alternative to serving
all queries with a single powerful model, enabling more favorable
quality--efficiency trade-offs at inference time.  Existing approaches
typically learn routing policies from query--model performance observations,
training a router to predict model quality for each
query \citep{zhang2025beyond,feng2025graphrouter,song2025irt}. Such supervision
is usually collected for a particular query workload and candidate model pool,
causing the resulting router to specialize to the routing environment in which
it is trained.

We term this environment-specific training paradigm \textbf{local fitting},
because the router is optimized from behavioral observations collected within
a particular routing environment and thereby specializes to its query
distribution and candidate model pool, rather than learning a routing capability
that is shared across environments. In practice, however, routing environments
are inherently dynamic. New domains and modalities emerge
\citep{ma2026mmr}, while candidate pools evolve as new models are introduced
and existing ones are replaced \citep{wang2026icl}. Consequently, when the
domain, candidate pool, or modality changes, conventional routers typically
require additional behavioral observations to be collected and the routing
model to be optimized again. This repeated fit-and-refit process increases the
cost of maintaining routing systems and, more fundamentally, makes the learned
router itself difficult to reuse across environments.

\begin{figure}[t]
	\centering
	\includegraphics[width=\linewidth]{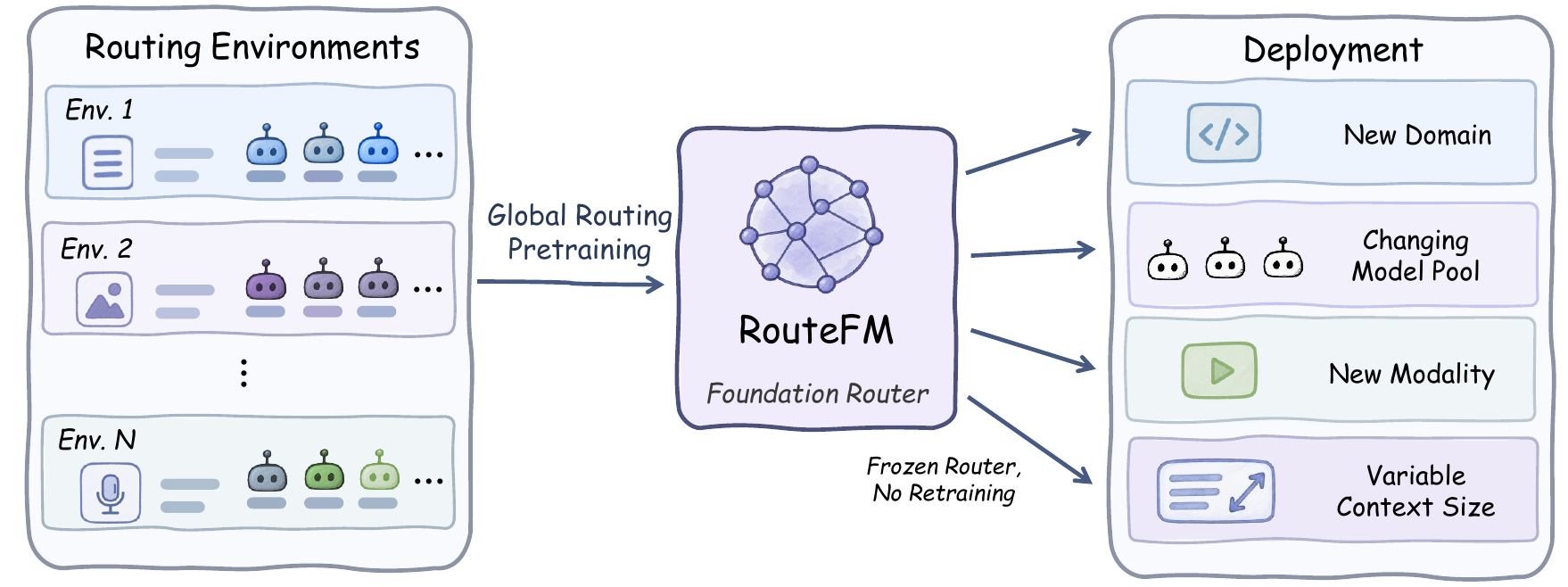}
	\caption{
    From local fitting to global routing pretraining.
    Traditional routers are optimized for individual routing environments, whereas RouteFM learns a reusable routing capability across heterogeneous environments and adapts to new environments through context.
    }
	\label{fig:intro}
    \vspace{-0.5cm}
\end{figure}

To move beyond this local-fitting paradigm, we ask a fundamental question:
\textbf{can LLM routing be approached from a foundation-model perspective, with
a single reusable router that generalizes across tasks, candidate
models, and deployment conditions?} At its core, routing is a target-conditioned
ranking problem: for a target query, the router must rank candidates by
suitability. Behavioral context provides evidence for this comparison, allowing
the router to infer which aspects of a candidate's behavior are relevant
to the target query. We argue that the ability to extract target-relevant
evidence and compare candidates can be shared across routing environments. In
practice, this shared capability should not depend on fixed model identities;
instead, candidates can be characterized by their observed behavior on prior
queries \citep{varangot2026generalising}. Based on this view, we
propose a \textbf{foundation-model paradigm for LLM routing}: a single routing
model learns this reusable target-conditioned comparison capability across
heterogeneous environments, while each deployment is specified by the
behavioral evidence available for its candidate pool.

Realizing this paradigm requires a router to learn a target-conditioned
comparison capability that transfers across environments, rather than a policy
tied to specific models or deployments. We introduce \textbf{RouteFM} to
address this challenge. As illustrated in Figure~\ref{fig:intro}, RouteFM
learns a reusable routing capability through global routing pretraining across
heterogeneous routing environments, and is subsequently deployed as a frozen
router under changing domains, candidate pools, modalities, and context sizes.
During pretraining, RouteFM is exposed to routing episodes that differ in
task, candidate-pool composition, and context size. Across these episodes, the
same underlying routing structure recurs: candidate capabilities are inferred
from observed behavior, target-relevant evidence is identified, and candidates
are compared accordingly. At inference time,
given a target query and behavioral observations of anonymous candidates,
RouteFM infers their capabilities in context and determines their suitability
for the query. The same pretrained router can therefore incorporate newly
introduced models and operate across new domains, modalities, and context
budgets without environment-specific retraining.

We evaluate RouteFM across in-domain routing, cross-modal transfer, and changing deployment conditions. A single frozen RouteFM performs strongly in-domain and further generalizes to MMR-Bench, an unseen multimodal routing benchmark excluded from pretraining, where it outperforms the strongest non-RouteFM baseline by 2.23 quality points with only eight behavioral observations per candidate. Beyond cross-modal transfer, RouteFM can incorporate newly introduced models from only a few observations, operate in new target domains without parameter optimization, and retain competitive routing quality under substantially reduced context budgets. These results show that a pretrained routing capability can be reused across changing routing environments primarily through contextual adaptation rather than repeated parameter optimization.

Our contributions are threefold:
\begin{itemize}[leftmargin=*]
    \item We introduce a \textbf{foundation-model paradigm for LLM routing}, moving beyond environment-specific local fitting toward a reusable routing capability that can generalize across changing tasks, candidate models, and deployment conditions.
    
    \item We propose \textbf{RouteFM}, combining global routing pretraining with in-context capability inference to adapt to new routing environments without parameter updates.
    
    \item We show that a single frozen RouteFM transfers across domains and modalities and incorporates newly introduced models from limited observations, all without retraining.
\end{itemize}
\section{Related Work}

\noindent \textbf{LLM Routing.}
LLM routing selects a suitable model for each query from a heterogeneous candidate pool, typically balancing response quality, inference cost, and other deployment objectives \citep{chen2024frugalgpt,ong2024routellm,feng2025graphrouter,song2025irt}. Existing methods include cascade-based selection, preference or performance prediction, and cost-aware routing \citep{aggarwal2024automix,ding2024hybrid,mei2025omnirouter,ding2025bestroute}. Recent studies further examine the reliability of routing itself, including supervision quality \citep{lai2026dars}, degenerate model-selection behaviors \citep{lai2026routing}, and evaluation under heterogeneous user preferences \citep{lai2026routejudge}. Meanwhile, benchmarks such as RouterBench, RouterEval, and MMR-Bench have expanded routing evaluation across broader tasks, models, and modalities \citep{hu2024routerbench,huang2025routereval,ma2026mmr}. Despite these advances, most routers are still trained or configured within a particular routing environment, following a local-fitting paradigm in which the learned router is tied to the task distribution and candidate pool from which its supervision is collected.

\noindent \textbf{Generalizable and Adaptive LLM Routing.}
Recent work has begun to relax the local-fitting assumption in LLM routing. IRT-Router improves cold-start generalization by explicitly modeling model capabilities and query characteristics \citep{song2025irt}, while ICL-Router derives model representations from in-context performance observations, allowing unseen models to be incorporated without retraining \citep{wang2026icl}. Retrieval-based approaches similarly estimate candidate performance from a small number of historical query--performance records and naturally support dynamic candidate pools \citep{varangot2026generalising}. These methods improve generalization or adaptation under particular deployment changes. RouteFM instead approaches LLM routing from a foundation-model perspective,
aiming to learn a reusable routing capability across heterogeneous
environments. It instantiates this paradigm through global routing
pretraining and in-context capability inference.
\section{Problem Formulation}
\label{sec:formulation}

We consider a collection of heterogeneous LLM routing environments
\(
\mathfrak{E}
=
\{\mathcal{E}_r=(\mathcal{D}_r,\mathcal{M}_r)\}_{r=1}^{R},
\)
where $\mathcal{D}_r$ denotes the query distribution in environment $r$ and
$\mathcal{M}_r=\{m_{r,1},\ldots,m_{r,M_r}\}$ is its candidate model pool.
Both the query distribution and candidate pool may vary across environments.

\subsection{Local Fitting in a Routing Environment}

Conventional routing methods typically fit a separate router within each
environment $\mathcal{E}_r$, yielding an environment-specific mapping
\[
f_{\theta_{\mathcal{E}_r}}: q \rightarrow m,
\qquad
q\sim\mathcal{D}_r,\quad m\in\mathcal{M}_r.
\]
The parameters $\theta_{\mathcal{E}_r}$ are optimized from query--model
performance observations collected in $\mathcal{E}_r$ and may therefore
specialize to its query distribution and candidate pool. When either
$\mathcal{D}_r$ or $\mathcal{M}_r$ changes, the learned routing function may
no longer transfer directly, and additional supervision or optimization can
be required.

\subsection{Routing as a Foundation Model}

We instead seek a routing capability that can be shared across environments.
Although routing environments may differ in their query distributions and
candidate model pools, they share the same underlying decision structure:
infer the capabilities of the available models and determine which is best
suited to the target query. This motivates learning \emph{how to route}
across environments rather than fitting an independent routing policy to each
one.

Rather than learning a separate $f_{\theta_{\mathcal{E}_r}}$ for every
$\mathcal{E}_r\in\mathfrak{E}$, we seek a single shared routing function
\[
f_{\theta}(q \mid \mathcal{E}_r) \rightarrow m_r^{*},
\qquad
q\sim\mathcal{D}_r,\quad
m_r^{*}\in\mathcal{M}_r,
\quad
\forall\,\mathcal{E}_r\in\mathfrak{E},
\]
where the same parameters $\theta$ are shared across routing environments.
Here, conditioning on $\mathcal{E}_r$ denotes adapting the routing decision
to the current query distribution and candidate pool, rather than learning a
new set of router parameters.

The goal is therefore to learn a reusable routing capability whose decisions
adapt as $\mathcal{D}_r$ and $\mathcal{M}_r$ vary, while the routing model
itself remains fixed. This defines the central objective of a
\textbf{foundation model for LLM routing}: a single shared model that can
operate across heterogeneous and evolving routing environments without
environment-specific retraining.
\section{RouteFM: A Foundation Model for LLM Routing}
\label{sec:routefm}

\begin{figure}[t]
    \centering
    \includegraphics[width=\linewidth]{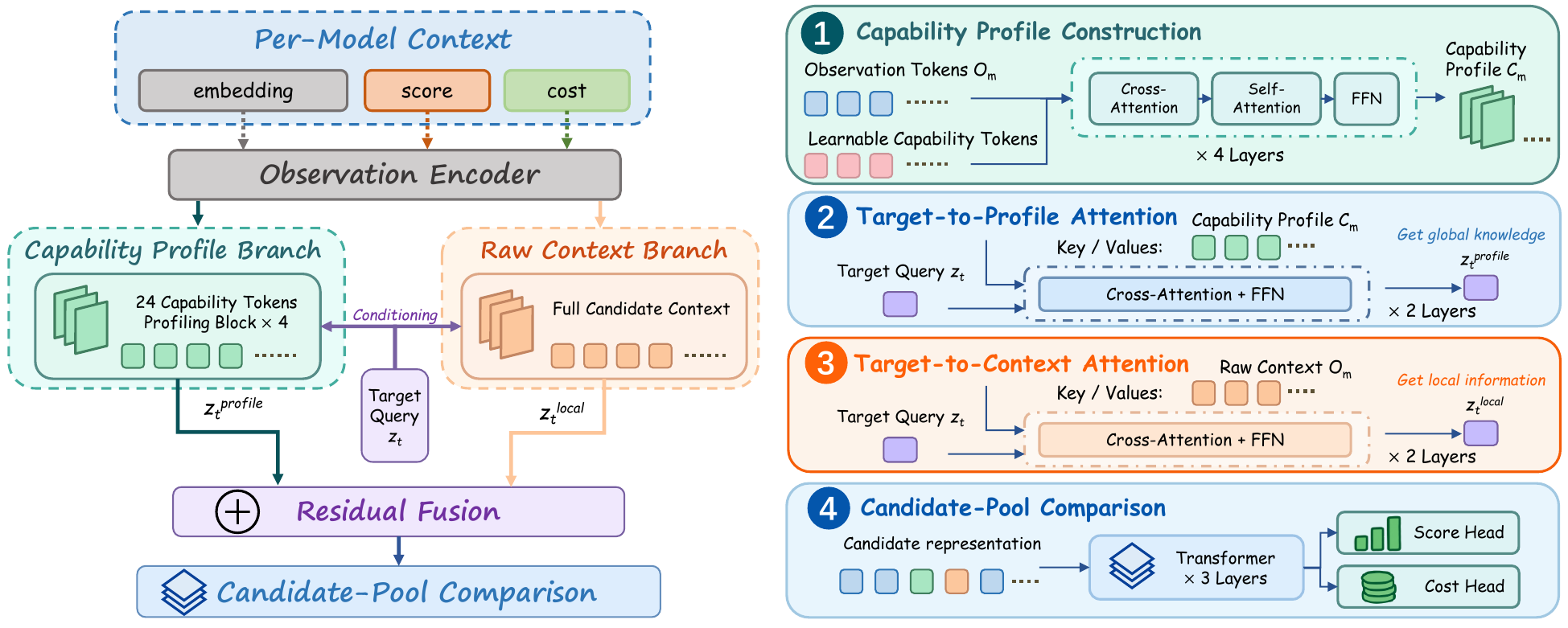}
    \caption{
    Architecture of RouteFM.
    Given behavioral observations for anonymous candidate models, RouteFM first
    constructs a compact capability profile for each candidate. The target query
    then retrieves complementary evidence from both the capability profile and
    the original behavioral context. These two representations are fused and
    jointly compared across the current candidate pool to predict target-specific
    quality and relative cost.
    }
    \label{fig:method}
\end{figure}

\subsection{Overview}

RouteFM implements the shared routing procedure described above over an
anonymous candidate pool specified by behavioral observations. Given a target
query and behavioral context for each candidate, RouteFM first encodes the
observations into a compact capability profile. As illustrated in
Figure~\ref{fig:method}, the target retrieves relevant evidence from both the
capability profile and the original context, after which RouteFM fuses the two
views and compares candidates within the current pool to predict target-specific
quality and relative cost. All candidate-specific information is inferred from
behavioral evidence; no model identities or persistent model embeddings are
provided.

\subsection{In-Context Capability Profiling}
\label{sec:capability}

A central challenge in reusable routing is that both query distributions and
candidate model pools vary across environments, making representations tied to
a fixed task or model identity difficult to reuse. RouteFM addresses this by
constructing each candidate representation directly from its observed behavior
in the current routing environment, without relying on a persistent identity.

For candidate model $m$, let
\[
\mathcal{B}_m
=
\{(e_i, y_{m,i}, \tilde c_{m,i})\}_{i=1}^{K_m},
\]
denote its behavioral context,
where $e_i\in\mathbb{R}^{4096}$ is the frozen embedding of a historical query,
$y_{m,i}\in[0,1]$ denotes the observed quality of model $m$ on that query, and
$\tilde c_{m,i}\in[0,1]$ denotes its normalized relative cost. Since the
routing corpora expose heterogeneous cost signals, we normalize cost within
each routing episode and use it only as a relative indicator. 

\noindent \textbf{Behavioral observation encoding.}
An observed outcome is informative only together with the query on which it
was obtained. For example, the same quality score may provide very different
evidence about a model when observed on queries requiring different
capabilities. RouteFM therefore encodes query semantics, observed quality, and
relative cost jointly rather than treating quality and cost as
query-independent model attributes.

Concretely, the query embedding is first projected into the router hidden
space, while quality and cost are mapped to learned outcome embeddings.
RouteFM further models their interaction with the query representation,
producing one behavioral token for each historical observation. We denote the
resulting sequence for candidate $m$ as
\[
O_m=[o_{m,1},\ldots,o_{m,K_m}]
\in\mathbb{R}^{K_m\times d}.
\]
The same behavioral encoder is shared by all candidates and routing
environments. By expressing query semantics and observed outcomes in a common
representation space, RouteFM can process behavioral evidence drawn from
different query distributions and candidate pools without introducing
environment-specific parameters. Thus, differences between candidate
representations arise entirely from their observed behavior, rather than from
candidate-specific parameters.

\noindent \textbf{Capability profile construction.}
A second challenge is that the amount of behavioral evidence available for a
candidate can vary substantially across environments and deployments. As shown
in Step~1 of Figure~\ref{fig:method}, RouteFM maps the variable-length
observation sequence $O_m$ into a fixed number of latent capability tokens.
RouteFM introduces a shared set of $L$ learnable latent capability tokens,
which initialize the profile of every candidate and become candidate-specific
only after interacting with its behavioral sequence $O_m$.

Each profiling block first lets the capability tokens retrieve information
from the behavioral observations through cross-attention, and then allows the
tokens to exchange information through self-attention. Stacking these blocks
yields
\(
C_m\in\mathbb{R}^{L\times d},
\)
which serves as the capability profile of candidate $m$. In our implementation,
$L=24$ and $d=256$.

Importantly, the capability profile is not supervised by predefined skill
categories. It is learned end-to-end from routing supervision, allowing the
latent tokens to capture behavioral factors that are useful for discriminating
among candidates across different queries and routing environments. RouteFM
therefore maintains two complementary representations for each candidate:
the compact capability profile $C_m$, which summarizes its overall behavior,
and the observation sequence $O_m$, which preserves fine-grained historical
evidence.

\subsection{Target-Conditioned Model Comparison}
\label{sec:comparison}

The capability profile summarizes a candidate's overall behavior, but routing
requires determining which evidence is relevant to the current target query.
Let $e_t$ denote the frozen embedding of target query $q_t$, and let
\[
z_t=P(e_t)\in\mathbb{R}^{d}
\]
be its projection into the router hidden space. For each candidate $m$,
RouteFM conditions the target representation on both its capability profile
$C_m$ and the original observation sequence $O_m$.

\noindent \textbf{Target-conditioned evidence retrieval.}
As illustrated in Figure~\ref{fig:method}, RouteFM retrieves complementary
evidence through two parallel attention branches:
\[
z_{t,m}^{p}
=
\mathrm{TargetAttend}(z_t,C_m),
\qquad
z_{t,m}^{c}
=
\mathrm{TargetAttend}(z_t,O_m).
\]
Both branches use two residual cross-attention blocks with the target
representation as the query. The profile branch captures how the target aligns
with the candidate's overall capabilities, whereas the context branch preserves
fine-grained evidence from individual historical observations that may be lost
during profile compression.

We combine the two sources through residual fusion,
\[
z_{t,m}
=
z_{t,m}^{p}
+
\alpha
\left(
z_{t,m}^{c}-z_t
\right),
\]
where $\alpha$ is a learned scalar shared across candidates. This design uses
the capability profile as the primary representation while allowing the raw
behavioral context to contribute additional target-specific evidence.

\noindent \textbf{Candidate-pool comparison.}
Routing is inherently comparative: the suitability of a candidate depends not
only on its own target-conditioned representation, but also on the alternatives
available in the current pool. RouteFM therefore jointly processes all candidate
representations,
\[
H_t^{(0)}
=
[z_{t,1},\ldots,z_{t,M}],
\]
using a Transformer encoder over the candidate dimension,
\[
H_t
=
\mathrm{Transformer}
\left(
H_t^{(0)}
\right).
\]
No positional encoding is added to candidate slots, making the comparison
permutation equivariant with respect to candidate ordering. Invalid or padded
candidates are masked throughout the computation.

Finally, independent prediction heads map each contextualized candidate
representation $H_{t,m}$ to its target-specific quality and relative cost,
\[
\hat y_{t,m}
=
\sigma\!\left(g_y(H_{t,m})\right),
\qquad
\hat c_{t,m}
=
\sigma\!\left(g_c(H_{t,m})\right).
\]
These predictions can be used directly for quality-based routing or combined
with deployment-specific preferences to construct downstream routing
objectives without updating RouteFM.

\subsection{Episodic Routing Pretraining}
\label{sec:pretraining}

To learn a routing procedure that transfers across environments, RouteFM is
pretrained episodically rather than on a fixed candidate pool. Each episode is constructed from a single task and contains an
anonymous candidate set $\mathcal{M}$, behavioral contexts
$\{\mathcal{B}_m\}_{m\in\mathcal{M}}$, and a disjoint set of target
queries $\mathcal{Q}$. Across episodes, we vary
the task, candidate composition, candidate ordering, and context size.
Candidate slots are randomly permuted, preventing RouteFM from associating
fixed positions with particular models and forcing it to infer candidate
capabilities from behavioral evidence.

\noindent \textbf{Quality and cost supervision.}
For each target query $q_t\in\mathcal{Q}$, RouteFM predicts both the quality
and relative cost of every valid candidate. For either quantity
$v\in\{y,c\}$, we combine point-wise regression with pairwise ranking,
\[
\mathcal{L}_v
=
\mathcal{L}^{\mathrm{point}}_v
+
\beta_v \mathcal{L}^{\mathrm{pair}}_v,
\]
where $\beta_v$ controls the contribution of the pairwise objective. The
point-wise term encourages accurate value prediction, while the pairwise term
preserves the relative ordering among candidates. Together, they supervise
both candidate capability estimation and the comparisons required for routing.

\noindent \textbf{Routing-aware regret.}
Accurate value prediction does not necessarily lead to the best routing
decision. We therefore additionally optimize a routing-regret objective. Given
the predicted qualities, we define a soft routing distribution
\[
\pi_{t,m}
=
\frac{\exp(\hat y_{t,m}/\tau_r)}
{\sum_{j\in\mathcal{M}_t}\exp(\hat y_{t,j}/\tau_r)},
\]
and minimize
\[
\mathcal{L}_{\mathrm{regret}}
=
\frac{1}{|\mathcal{Q}|}
\sum_t
\left[
\max_{m\in\mathcal{M}_t} y_{t,m}
-
\sum_{m\in\mathcal{M}_t}
\pi_{t,m}y_{t,m}
\right].
\]

The overall pretraining objective is
\[
\mathcal{L}
=
\lambda_y\mathcal{L}_y
+
\lambda_c\mathcal{L}_c
+
\lambda_r\mathcal{L}_{\mathrm{regret}},
\]
where $\lambda_y$, $\lambda_c$, and $\lambda_r$ balance the three objectives,
with the regret contribution progressively increased over the pretraining
curriculum. Detailed loss definitions and hyperparameter settings are provided
in Appendix~\ref{app:training}.

\subsection{Routing decision}
For each candidate model $m\in\mathcal{M}$, RouteFM predicts its
target-specific quality $\hat y_m$ and relative cost $\hat c_m$. The final
routing decision is made by maximizing a deployment-specific utility,
\[
m^{*}
=
\arg\max_{m\in\mathcal{M}}
\left(
\hat y_m-\lambda \hat c_m
\right),
\]
where $\lambda\ge 0$ controls the trade-off between response quality and
inference cost. Varying $\lambda$ allows the same pretrained router to support
different deployment preferences without retraining.
\section{Experiments}
\subsection{Experimental Setup}

\noindent \textbf{Pretraining.}
We pretrain RouteFM using four data sources: LLMRouterBench \citep{li2026llmrouterbench},
RouterBench \citep{hu2024routerbench}, RouterEval
\citep{huang2025routereval}, and MixInstruct \citep{jiang2023llm}. After
preprocessing, they contain 212,470 query--task instances and 1.73M valid
query--model observations.  Queries are represented by frozen
4,096-dimensional Qwen3-VL embeddings \citep{Qwen3-VL}, while model names, providers,
parameter counts, and other explicit identity features are never exposed
to RouteFM.

RouteFM itself is trained from random initialization using episodic
pretraining. Each episode contains $6$--$24$ anonymous candidates,
$16$--$32$ target queries disjoint from the behavioral context, and a
per-model observation budget
$K\in\{8,16,32,64,128,256,512,1024\}$.
Across episodes, we vary the task, candidate-pool composition, candidate
ordering, and observation budget. Training runs for 10,000 AdamW steps with an effective
batch size of 16 and follows a curriculum from large observation sets toward
deployment-scale regimes. Full preprocessing, episode
construction, curriculum, and optimization details are provided in
Appendix~\ref{app:training}.

\noindent \textbf{Evaluation environments.}
We first evaluate in-domain routing on RouterEval
\citep{huang2025routereval}, using held-out queries from tasks included in
pretraining. Each candidate pool contains four anonymous models, and we report
results under limited behavioral context with
$K\in\{8,16,32,64\}$ observations per candidate.
We then evaluate transfer to MMR-Bench \citep{ma2026mmr}, which is excluded
from pretraining and provides a substantially different multimodal routing
environment. In the limited-observation regime, we evaluate
$K\in\{8,16,32,64\}$ using dataset-wise five-fold splits that separate the
behavioral context from evaluation queries. We additionally consider a large-observation regime in which $40\%$ of each
dataset is used as behavioral context and the remaining $60\%$ is reserved
for evaluation.
RouteFM remains frozen throughout evaluation and never uses evaluation labels
for parameter updates.

\noindent \textbf{Baselines and metric.}
We compare RouteFM with a broad set of routing baselines, including
Weighted $k$NN \citep{stripelis2024tensoropera},
MLP \citep{stripelis2024tensoropera},
Embedding-Ridge,
EmbedLLM \citep{zhuang2025embedllm},
RM-Softmax \citep{tsiourvas2025causal},
TRouter \citep{liu2026task},
InferenceDynamics \citep{shi2025inference},
UniRoute \citep{jitkrittum2026universal}, ICL-Router \citep{wang2026icl}, 
and Context-Mean.
All baselines are implemented following the unified reproduction framework of
\citet{lai2026orbit}, using matched candidate pools, behavioral observations,
evaluation queries, and query representations.
Our primary metric is the average realized quality of the selected model.
Unless otherwise specified, routing decisions select the candidate with the
highest predicted quality.
Further implementation and evaluation details are provided in
Appendix~\ref{app:evaluation}.

\subsection{Main Result}

\noindent \textbf{In-Domain Routing.}
We first evaluate RouteFM on held-out RouterEval queries from tasks represented
during pretraining. As shown in Table~\ref{tab:adaptation_main}, RouteFM
achieves the highest routing quality across all observation budgets. Its
advantage is largest at $K=8$, where it outperforms the strongest baseline by
$0.84$ quality points, while the margins become smaller as more behavioral
evidence is available. This trend suggests that the pretrained router can make
effective use of sparse behavioral evidence, whereas the advantage naturally
shrinks once all methods are given richer observations about the candidate
models. Importantly, the same frozen RouteFM remains competitive across the
entire context range, indicating that cross-environment pretraining does not
come at the expense of in-domain routing quality.

\noindent \textbf{Cross-Modal Transfer.}
We next evaluate whether the routing capability learned by RouteFM transfers
to a multimodal routing environment excluded from pretraining. As shown in
Table~\ref{tab:adaptation_main}, RouteFM achieves the highest routing quality
across all observation budgets on MMR-Bench. The advantage is most pronounced
at $K=8$, where RouteFM outperforms the strongest non-RouteFM baseline by
$2.23$ quality points. As more behavioral evidence becomes available, the
margin generally decreases, reaching only $0.07$ points in the large-observation
regime. This setting introduces a stronger distribution shift than the
in-domain evaluation, since both the query characteristics and the modality
differ from those seen during pretraining. Nevertheless, a small amount of
behavioral context is sufficient for the frozen RouteFM to adapt to the new
candidate behavior without parameter updates. These results show that
RouteFM's advantage is concentrated in the low-context regime, indicating
strong observation efficiency when transferring to a new routing environment.

\begin{table*}[t]
\centering
\caption{
Routing quality under in-domain routing and cross-modal transfer.
All methods use matched candidate pools and behavioral observation budgets
within each setting. ``Large'' denotes the large-observation regime using
$40\%$ of each MMR-Bench dataset as behavioral context.
Best results are bolded and second-best results are underlined.
}
\label{tab:adaptation_main}

\resizebox{\textwidth}{!}{
\begin{tabular}{@{}l*{9}{c}@{}}
\toprule
\multirow{2}{*}{Method}
& \multicolumn{4}{c}{In-Domain Routing}
& \multicolumn{5}{c}{Cross-Modal Transfer}  \\
\cmidrule(lr){2-5}
\cmidrule(l){6-10}
& $K{=}8$ & $K{=}16$ & $K{=}32$ & $K{=}64$
& $K{=}8$ & $K{=}16$ & $K{=}32$ & $K{=}64$ & Large \\
\midrule

Weighted $k$NN
& 0.6066 & 0.6273 & 0.6390 & 0.6406
& 0.6975 & 0.7123 & 0.7159 & 0.7193 & 0.7282 \\

MLP
& 0.6021 & 0.6351 & 0.6561 & 0.6578
& 0.7026 & 0.7283 & 0.7333 & 0.7400 & 0.7523 \\

Embedding-Ridge
& \underline{0.6108} & 0.6348 & 0.6568 & 0.6520
& 0.7058 & 0.7293 & 0.7391 & 0.7408 & 0.7517 \\

EmbedLLM
& 0.6019 & \underline{0.6430} & 0.6570 & \underline{0.6601}
& \underline{0.7100} & \underline{0.7338} & 0.7374 & 0.7380 & 0.7604 \\

RM-Softmax
& 0.6101 & 0.6390 & \underline{0.6601} & 0.6586
& 0.7021 & 0.7285 & 0.7405 & 0.7433 & 0.7604 \\

TRouter
& 0.6064 & 0.6341 & 0.6559 & 0.6568
& 0.7067 & 0.7311 & \underline{0.7433} & \underline{0.7440} & 0.7576 \\

InferenceDyn.
& 0.6064 & 0.6341 & 0.6559 & 0.6568
& 0.7067 & 0.7311 & \underline{0.7433} & \underline{0.7440} & 0.7576 \\

UniRoute
& 0.6092 & 0.6413 & 0.6548 & 0.6586
& 0.6990 & 0.7263 & 0.7357 & 0.7431 & \underline{0.7607} \\

ICL-Router
& 0.4506 & 0.4955 & 0.4926 & 0.4920
& 0.7054 & 0.7185 & 0.7182 & 0.7327 & 0.7448 \\

Context-Mean
& 0.6064 & 0.6341 & 0.6559 & 0.6568
& 0.7067 & 0.7311 & \underline{0.7433} & \underline{0.7440} & 0.7576 \\
\midrule

\textbf{RouteFM}
& \textbf{0.6192}
& \textbf{0.6433}
& \textbf{0.6627}
& \textbf{0.6611}
& \textbf{0.7323}
& \textbf{0.7457}
& \textbf{0.7487}
& \textbf{0.7504}
& \textbf{0.7614} \\

\bottomrule
\end{tabular}
}
\vspace{-0.1cm}
\end{table*}

\subsection{Ablation Study}

\noindent \textbf{Architecture.}
Table~\ref{tab:mmrbench_ablation}(a) examines the main architectural
components of RouteFM. Removing the candidate-pool Transformer causes the
largest degradation, with a $1.33$-point drop at $K=8$, indicating the
importance of jointly comparing candidates rather than estimating them
independently. Removing target-to-context attention also consistently reduces
performance, particularly under limited observations, showing that direct
access to individual behavioral observations provides complementary evidence
beyond the compressed capability profile. Replacing the Qwen3-VL query
representation with a BGE encoder further degrades performance across
observation budgets, indicating that the underlying query representation also
contributes to effective cross-modal routing.

\noindent \textbf{Pretraining and target-domain adaptation.}
Table~\ref{tab:mmrbench_ablation}(b) uses a separate support-controlled
protocol to study the role of routing pretraining and target-domain
optimization. The training-from-scratch and fine-tuning variants are given
$20\%$ labeled target-domain support data for parameter optimization, whereas
the frozen RouteFM uses no target-domain labels for parameter updates.
Training the same architecture from scratch consistently underperforms the
frozen RouteFM, with a $1.08$-point gap at $K=8$. Starting from the pretrained
RouteFM and further fine-tuning on the target domain yields only small changes:
it provides marginal gains at some observation budgets and slightly lower
performance at others. These results indicate that much of the transferable
routing capability is already acquired during pretraining, allowing RouteFM to
adapt to the target environment through behavioral context without additional
parameter optimization.

\begin{table*}[t]
\centering 
\caption{
Ablations on MMR-Bench.
Part~(a) studies architectural components under the standard evaluation
protocol. Part~(b) follows a separate support-controlled protocol in which
target-domain training and fine-tuning use $20\%$ labeled support data, while
frozen RouteFM performs no target-domain parameter updates.
}
\label{tab:mmrbench_ablation}

\small
\setlength{\tabcolsep}{8pt}
\begin{tabular}{@{}lccccc@{}}
\toprule
\multirow{2}{*}{Method}
& \multicolumn{4}{c}{Limited Observations}
& \multicolumn{1}{c}{Large Budget} \\
\cmidrule(lr){2-5}
\cmidrule(l){6-6}
& $K{=}8$ & $K{=}16$ & $K{=}32$ & $K{=}64$ & Large \\
\midrule

\multicolumn{6}{@{}l}{\textit{(a) Architecture}} \\[1pt]

\textbf{RouteFM}
& \textbf{0.7323}
& \textbf{0.7457}
& \textbf{0.7487}
& \textbf{0.7504}
& \textbf{0.7614} \\

w/o target-to-context attention
& 0.7266
& 0.7407
& 0.7475
& \textbf{0.7504}
& 0.7568 \\

w/o candidate-pool Transformer
& 0.7190
& 0.7380
& 0.7403
& 0.7433
& 0.7477 \\

BGE text-only encoder
& 0.7308
& 0.7384
& 0.7468
& 0.7475
& 0.7491 \\

\midrule

\multicolumn{6}{@{}l}{\textit{(b) Pretraining and target-domain adaptation}} \\[1pt]

\textbf{RouteFM (frozen)}
& \textbf{0.7523}
& 0.7534
& 0.7538
& \textbf{0.7537}
& \textbf{0.7584} \\

Target-domain training from scratch
& 0.7415
& 0.7495
& 0.7523
& 0.7521
& 0.7525 \\

RouteFM + target-domain fine-tuning
& 0.7510
& \textbf{0.7535}
& \textbf{0.7543}
& 0.7534
& 0.7575 \\

\bottomrule
\end{tabular}
\end{table*}

\subsection{Adaptation to Deployment Changes}

%Figure~\ref{fig:deployment_adaptation} examines how a frozen RouteFM adapts when different aspects of the routing environment change.

\begin{figure*}[t]
    \centering
    \begin{minipage}[t]{0.32\textwidth}
        \centering
        \includegraphics[width=\linewidth]{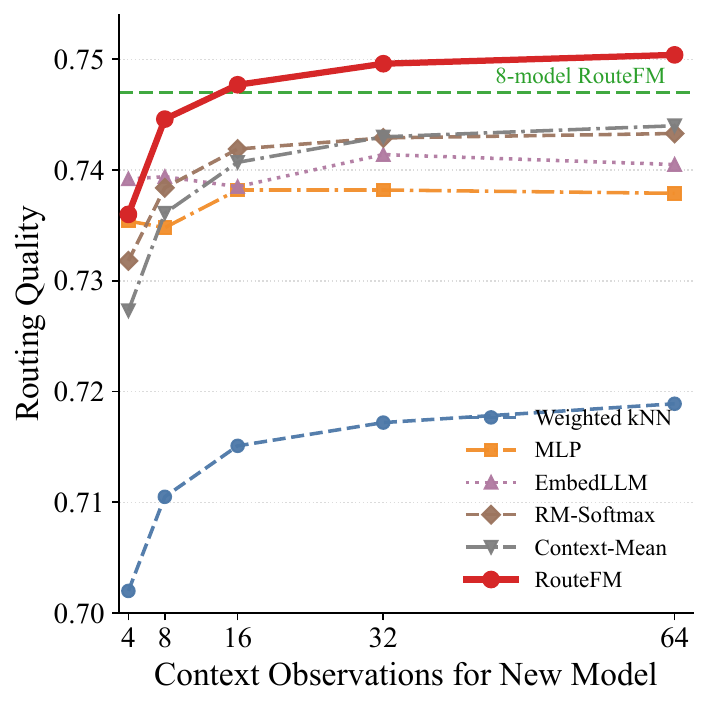}
        \vspace{-2mm}
        \small \textbf{(a) New-model incorporation}
    \end{minipage}
    \hfill
    \begin{minipage}[t]{0.32\textwidth}
        \centering
        \includegraphics[width=\linewidth]{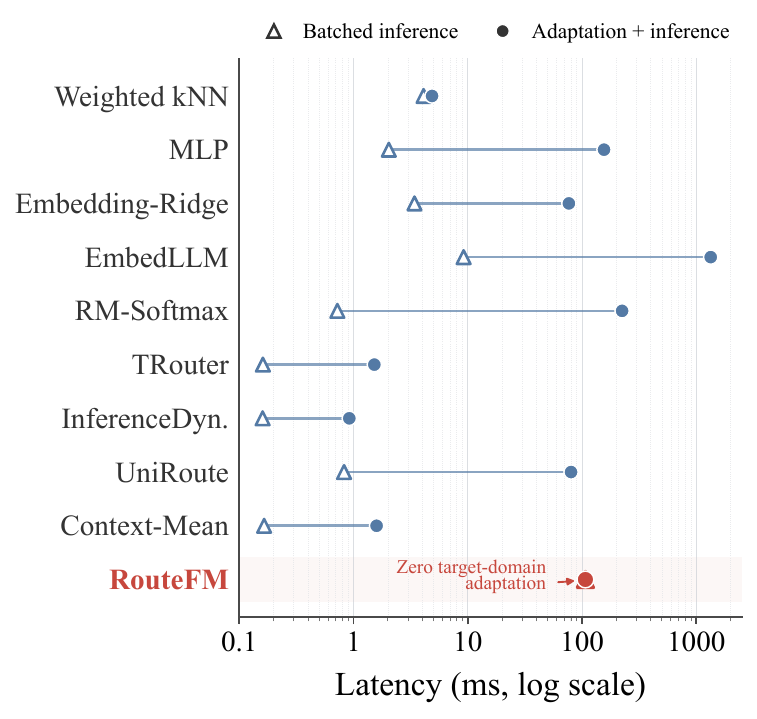}
        \vspace{-2mm}
        \small \textbf{(b) Target-domain efficiency}
    \end{minipage}
    \hfill
    \begin{minipage}[t]{0.32\textwidth}
        \centering
        \includegraphics[width=\linewidth]{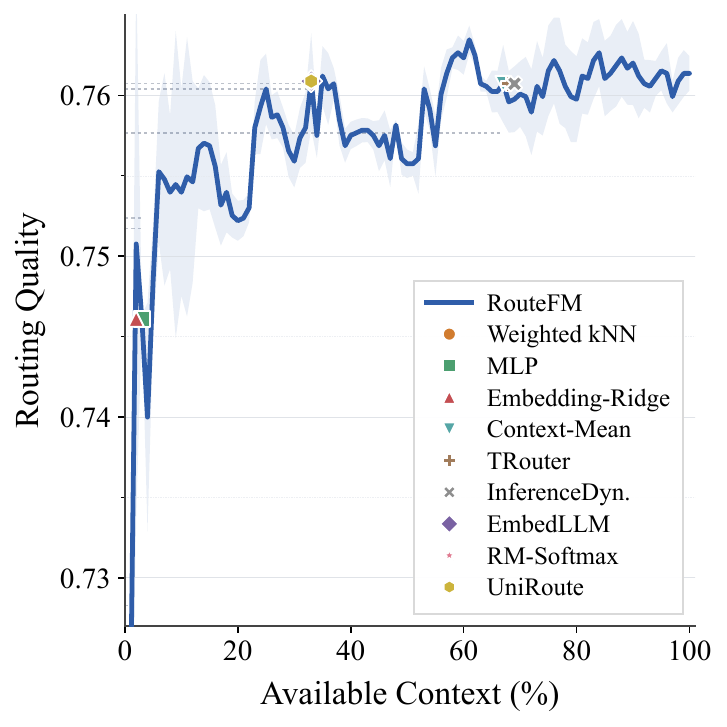}
        \vspace{-2mm}
        \small \textbf{(c) Context-budget efficiency}
    \end{minipage}
\caption{
Adaptation of frozen RouteFM to deployment changes:
\textbf{(a)} new-model incorporation,
\textbf{(b)} target-domain efficiency, and
\textbf{(c)} context-budget efficiency.
Markers in \textbf{(c)} denote the context fraction required to match each
baseline's full-context routing quality.
}
\vspace{-0.3cm}
\label{fig:deployment_adaptation}
\end{figure*}

\noindent \textbf{New-model incorporation.}
We first study whether RouteFM can incorporate a newly introduced model without
retraining. Starting from an incumbent candidate pool, we add a newly introduced candidate and gradually increase the amount of behavioral evidence available for it. As shown in Figure~\ref{fig:deployment_adaptation}(a), RouteFM
surpasses all refitted baselines after observing only eight examples from the
new candidate. With 16 observations, routing over the expanded pool also
outperforms routing over the original incumbent pool, indicating that RouteFM
can not only characterize the new model from limited evidence but also exploit
it when it becomes useful. This demonstrates that candidate pools can be
expanded through contextual evidence without modifying the pretrained router.

\noindent \textbf{Target-domain efficiency.}
We next examine the overhead of operating in a new target domain. RouteFM is
applied directly with frozen parameters using the available behavioral context,
whereas learned environment-specific baselines require an additional fitting
stage before inference. Figure~\ref{fig:deployment_adaptation}(b) reports the
combined adaptation and inference cost. Although simple non-parametric methods
remain faster in absolute latency, RouteFM avoids target-domain parameter
optimization while retaining the predictive capacity of a learned router.
This makes adaptation primarily a context-construction problem rather than a
retraining problem.

\noindent \textbf{Context-budget efficiency.}
Finally, we vary the amount of behavioral context available to RouteFM and
measure how much is required to recover the full-context performance of
competing routers. Figure~\ref{fig:deployment_adaptation}(c) uses a separate
dense context-budget protocol, with each competing router evaluated using its
full available context as a reference. RouteFM matches the full-context
performance of Weighted $k$NN with only $1\%$ of the available observations,
MLP with $3\%$, and EmbedLLM with $26\%$, while progressively larger context
fractions are required to match the stronger full-context references. The
shaded region shows variation across the three evaluation splits. Overall,
RouteFM retains competitive routing quality under substantially reduced
context budgets, showing that the pretrained routing capability can reduce the
amount of behavioral evidence required at deployment.

\section{Conclusion}

We introduce a \textbf{foundation-model paradigm for LLM routing}, shifting
the field from environment-specific local fitting toward reusable routing
models that can generalize across heterogeneous deployments. Rather than
rebuilding a routing policy for each new setting, this paradigm views routing
as a transferable capability that can support evolving tasks, candidate model
pools, and deployment conditions. We hope this perspective will motivate
future research on general-purpose routing models as reusable infrastructure
for LLM systems, advancing the vision of \emph{pretrain once, route anywhere}.

%\subsubsection*{Author Contributions}

%\subsubsection*{Acknowledgments}

\bibliography{iclr2027_conference}
\bibliographystyle{iclr2027_conference}

\appendix
\clearpage
\appendix

\setcounter{figure}{0}
\setcounter{table}{0}
\renewcommand{\thefigure}{A\arabic{figure}}
\renewcommand{\thetable}{A\arabic{table}}

\section*{Appendix}

\section{Architecture and Implementation Details}
\label{app:architecture}

This section provides the full architecture of RouteFM. Unless otherwise
specified, the router hidden dimension is $d=256$. All candidate-specific
representations are constructed from behavioral observations; model names,
provider identities, parameter counts, and persistent model-ID embeddings are
never provided to RouteFM.

\subsection{Inputs and Masking}

For each candidate model $m$, RouteFM receives a behavioral context
\[
\mathcal{B}_m
=
\{(e_{m,i}, y_{m,i}, \tilde c_{m,i})\}_{i=1}^{K_m}.
\]
where $e_{m,i}\in\mathbb{R}^{D}$ is the frozen embedding of a historical
query, $y_{m,i}\in[0,1]$ is the observed response quality, and
$\tilde c_{m,i}\in[0,1]$ is the normalized relative cost.

For a batch containing at most $M$ candidates, $K$ context positions, and
$T$ target queries, the main inputs are
\[
E^{\mathrm{ctx}}\in\mathbb{R}^{B\times M\times K\times D},
\qquad
F^{\mathrm{ctx}}\in\mathbb{R}^{B\times M\times K\times 2},
\]
and
\[
E^{\mathrm{tgt}}\in\mathbb{R}^{B\times T\times D}.
\]
The two context features correspond to observed quality and normalized cost.
Binary masks specify valid context observations, candidates, and
target--candidate pairs. Padding and missing observations are excluded from
all attention and loss computations.

Different target queries are processed independently. In particular, the
outcome of one target query is never available as behavioral evidence for
another target query.

The main RouteFM model uses $D=4096$ dimensional, L2-normalized Qwen3-VL
embeddings. The text-only BGE control uses $D=768$ dimensional
BGE-base-en-v1.5 representations. The two variants use the same routing
architecture after the input projection and are trained separately.

\subsection{Behavioral Observation Encoder}

The same query projector is applied to both behavioral-context queries and
target queries. For an input embedding $e$, we compute
\[
P(e)
=
W_2\,
\mathrm{Dropout}
\left(
\mathrm{GELU}
\left(
W_1\mathrm{LN}(e)+b_1
\right)
\right)
+b_2,
\]
where the intermediate dimension is $512$, the output dimension is
$d=256$, and the dropout probability is $0.1$.

For each behavioral observation, the scalar quality and cost signals are
independently embedded as
\[
s_{m,i}
=
\mathrm{MLP}_{y}(y_{m,i}),
\qquad
r_{m,i}
=
\mathrm{MLP}_{c}(\tilde c_{m,i}),
\]
where both scalar networks have architecture
\[
1\rightarrow256\rightarrow256.
\]

Simply adding these outcome embeddings to the query representation would
treat the same observed score similarly across semantically different
queries. RouteFM therefore explicitly models interactions between query
semantics and observed outcomes. Let
\[
q_{m,i}=P(e_{m,i}).
\]
We compute
\[
h_{m,i}
=
W_I
\left[
q_{m,i}\odot s_{m,i}
\ ;\
q_{m,i}\odot r_{m,i}
\right]
+b_I,
\]
where $\odot$ denotes element-wise multiplication and $[\cdot\,;\cdot]$
denotes concatenation.

The intermediate behavioral representation is
\[
b_{m,i}
=
q_{m,i}
+
s_{m,i}
+
r_{m,i}
+
h_{m,i}.
\]
A residual feed-forward transformation then produces the final observation
token,
\[
o_{m,i}
=
\mathrm{LN}
\left(
b_{m,i}
+
\mathrm{FFN}_{\mathrm{obs}}(b_{m,i})
\right),
\]
where the feed-forward hidden width is $1024$.

Collecting all valid observations of candidate $m$ gives
\[
O_m
=
[o_{m,1},\ldots,o_{m,K_m}]
\in
\mathbb{R}^{K_m\times d}.
\]

The behavioral encoder is shared by all candidates. Consequently, two
candidates can obtain different representations only through differences in
their observed query--quality--cost histories. The released architecture uses
exactly two behavioral outcome features, quality and relative cost; no
uncertainty feature or candidate-specific learned embedding is used.

\subsection{Capability Profile Construction}

The number of behavioral observations may vary substantially across
candidates and environments. RouteFM therefore compresses the variable-length
sequence $O_m$ into a fixed-capacity latent capability profile.

Each candidate starts from the same set of $L=24$ learnable capability tokens,
\[
C_m^{(0)}
=
C^{(0)}
\in
\mathbb{R}^{L\times d}.
\]
These tokens contain no candidate-specific information before observing
$O_m$.

RouteFM applies four capability-profiling blocks. Let
$C_m^{(\ell-1)}$ denote the capability tokens entering block $\ell$.
The block first retrieves information from the candidate's behavioral
observations through cross-attention,
\[
A_m^{(\ell)}
=
C_m^{(\ell-1)}
+
\mathrm{MHA}_{\mathrm{cross}}
\left(
\mathrm{LN}(C_m^{(\ell-1)}),
O_m,
O_m;
\mathcal{M}_m^{\mathrm{ctx}}
\right),
\]
where the capability tokens provide the queries, the observation tokens
provide keys and values, and $\mathcal{M}_m^{\mathrm{ctx}}$ masks invalid
context positions.

The capability tokens then exchange information through self-attention,
\[
B_m^{(\ell)}
=
A_m^{(\ell)}
+
\mathrm{MHA}_{\mathrm{self}}
\left(
\mathrm{LN}(A_m^{(\ell)}),
\mathrm{LN}(A_m^{(\ell)}),
\mathrm{LN}(A_m^{(\ell)})
\right).
\]
Finally, a position-wise feed-forward network updates each capability token,
\[
C_m^{(\ell)}
=
B_m^{(\ell)}
+
\mathrm{FFN}_{\mathrm{prof}}
\left(
\mathrm{LN}(B_m^{(\ell)})
\right).
\]

All profiling attention modules use eight heads, and the feed-forward hidden
width is $1024$. Residual connections and pre-normalization are used
throughout.

After four blocks,
\[
C_m
=
C_m^{(4)}
\in
\mathbb{R}^{24\times256}
\]
is used as the capability profile of candidate $m$.

The profile tokens are not assigned predefined meanings such as mathematics,
coding, or visual reasoning. Their roles emerge end-to-end from routing
supervision. The resulting $C_m$ therefore acts as a learned latent summary
of those behavioral characteristics that are useful for distinguishing
candidates across routing environments.

RouteFM retains both
\[
C_m
\qquad\text{and}\qquad
O_m.
\]
The former provides a compact global summary, while the latter preserves the
individual behavioral observations for later target-conditioned retrieval.

\subsection{Target-Conditioned Evidence Retrieval}

Capability profiling describes what a candidate has demonstrated globally,
but routing requires determining which evidence is relevant to a particular
target query.

For target query $q_t$, let
\[
z_t^{(0)}
=
P(e_t)
\in
\mathbb{R}^{d}.
\]
For each candidate, RouteFM applies two parallel target-conditioned retrieval
branches: one over the capability profile $C_m$ and one over the original
observation sequence $O_m$.

\noindent \textbf{Target-to-profile branch.}

The profile branch initializes its query state with
\[
z_{t,m}^{p,(0)}=z_t^{(0)}.
\]
It then applies two residual cross-attention blocks. For
$\ell\in\{1,2\}$,
\[
u_{t,m}^{p,(\ell)}
=
z_{t,m}^{p,(\ell-1)}
+
\mathrm{MHA}_{p}
\left(
\mathrm{LN}(z_{t,m}^{p,(\ell-1)}),
C_m,
C_m
\right),
\]
followed by
\[
z_{t,m}^{p,(\ell)}
=
u_{t,m}^{p,(\ell)}
+
\mathrm{FFN}_{p}
\left(
\mathrm{LN}(u_{t,m}^{p,(\ell)})
\right).
\]
The output of the second block is
\[
z_{t,m}^{p}
=
z_{t,m}^{p,(2)}.
\]

Because the target representation provides the attention query, this branch
does not merely read a generic summary of candidate $m$. Instead, it extracts
the part of the learned capability profile that is most relevant to the
current target query.

\noindent \textbf{Target-to-context branch.}

In parallel, the context branch starts from the same target representation,
\[
z_{t,m}^{c,(0)}=z_t^{(0)},
\]
but retrieves directly from the uncompressed observation sequence $O_m$.

For $\ell\in\{1,2\}$,
\[
u_{t,m}^{c,(\ell)}
=
z_{t,m}^{c,(\ell-1)}
+
\mathrm{MHA}_{c}
\left(
\mathrm{LN}(z_{t,m}^{c,(\ell-1)}),
O_m,
O_m;
\mathcal{M}_m^{\mathrm{ctx}}
\right),
\]
and
\[
z_{t,m}^{c,(\ell)}
=
u_{t,m}^{c,(\ell)}
+
\mathrm{FFN}_{c}
\left(
\mathrm{LN}(u_{t,m}^{c,(\ell)})
\right).
\]
The resulting local representation is
\[
z_{t,m}^{c}
=
z_{t,m}^{c,(2)}.
\]

This branch attends to all valid behavioral observations. RouteFM does not
apply a nearest-neighbor prefilter or introduce an externally specified
query-similarity bias before attention. Instead, the attention mechanism
itself learns which historical observations are relevant to the target.

The two branches therefore provide complementary views. The profile branch
retrieves target-relevant information from a compressed representation of the
candidate's overall behavior, whereas the context branch can recover
fine-grained evidence from individual historical observations.

\subsection{Residual Fusion}

Both target-conditioned branches are initialized from the same projected
target $z_t^{(0)}$. Because the context branch is residual, its output can be
viewed as the original target representation plus a context-induced update.
RouteFM therefore isolates this update as
\[
\Delta z_{t,m}^{c}
=
z_{t,m}^{c}
-
z_t^{(0)}.
\]

The final target-conditioned candidate representation is
\[
z_{t,m}
=
z_{t,m}^{p}
+
\alpha
\Delta z_{t,m}^{c}
=
z_{t,m}^{p}
+
\alpha
\left(
z_{t,m}^{c}
-
z_t^{(0)}
\right),
\]
where $\alpha$ is a learnable scalar shared across all candidates and target
queries and is initialized to $0.1$.

Subtracting the original target representation prevents the target residual
from being added twice. The profile branch therefore serves as the primary
target-conditioned candidate representation, while the context branch
contributes only the additional evidence induced by direct access to the raw
behavioral observations.

\subsection{Candidate-Pool Comparison}

After target-conditioned retrieval, RouteFM has one representation
$z_{t,m}$ for each candidate. Predicting each candidate independently,
however, would ignore the fact that routing is a relative decision. RouteFM
therefore jointly contextualizes all candidate representations before
prediction.

For target $t$, we form
\[
H_t^{(0)}
=
[z_{t,1},\ldots,z_{t,M}]
\in
\mathbb{R}^{M\times d}.
\]
A three-layer Transformer encoder operates along the candidate dimension.
For layer $\ell\in\{1,2,3\}$,
\[
A_t^{(\ell)}
=
H_t^{(\ell-1)}
+
\mathrm{MHA}_{\mathrm{cand}}
\left(
\mathrm{LN}(H_t^{(\ell-1)}),
\mathrm{LN}(H_t^{(\ell-1)}),
\mathrm{LN}(H_t^{(\ell-1)});
\mathcal{M}^{\mathrm{cand}}
\right),
\]
followed by
\[
H_t^{(\ell)}
=
A_t^{(\ell)}
+
\mathrm{FFN}_{\mathrm{cand}}
\left(
\mathrm{LN}(A_t^{(\ell)})
\right).
\]

Each layer uses hidden width $256$, eight attention heads, feed-forward width
$1024$, GELU activation, dropout $0.1$, residual connections, and
pre-normalization.

No positional encoding is added along the candidate dimension. Therefore,
for any permutation of the candidate ordering, the output representations are
permuted in the same way. Candidate comparison is consequently permutation
equivariant rather than tied to fixed candidate slots. Invalid or padded
candidates are masked from both self-attention and the final routing decision.

The output of the third layer is
\[
H_t
=
H_t^{(3)}
=
[H_{t,1},\ldots,H_{t,M}].
\]

\subsection{Prediction Heads and Routing}

Two independent prediction heads operate on each contextualized candidate
representation:
\[
\hat y_{t,m}
=
\sigma
\left(
w_y^\top H_{t,m}+b_y
\right),
\]
and
\[
\hat c_{t,m}
=
\sigma
\left(
w_c^\top H_{t,m}+b_c
\right).
\]

The quality head estimates target-specific response quality, whereas the cost
head predicts episode-relative cost. Under the default quality-only policy,
RouteFM selects
\[
m_t^*
=
\arg\max_{m\in\mathcal{M}_t}
\hat y_{t,m}.
\]

Because quality and cost are predicted separately, downstream deployment
objectives may alternatively combine them according to a desired
quality--cost preference without changing the RouteFM parameters.

The Qwen-based RouteFM contains $12.78$M trainable parameters, excluding the
frozen embedding encoder.

\subsection{Cost Preprocessing}

The pretraining sources expose heterogeneous cost signals, and these signals
do not necessarily correspond to the same physical quantity. We therefore
treat cost as an episode-relative feature rather than as an absolute dollar
price or latency measurement.

For raw cost $c$, we first apply a logarithmic transformation,
\[
c'
=
\log(1+c).
\]
Let $c'_{\min}$ and $c'_{\max}$ denote the minimum and maximum transformed
costs among valid behavioral-context observations in the current episode.
The normalized cost is
\[
\tilde c
=
\frac{
c'-c'_{\min}
}{
c'_{\max}-c'_{\min}
}.
\]
If $c'_{\max}=c'_{\min}$, all normalized context costs are set to zero.

Target cost labels used during training are transformed using the same
context-derived minimum and maximum and are clipped to $[0,1]$. Thus, target
information is not used to determine the normalization statistics.

RouterBench provides model parameter count as its available cost proxy.
Missing cost observations in other sources are represented by zero before
episode normalization. Consequently, the predicted
$\hat c_{t,m}$ should be interpreted as a relative cost signal within the
current episode and should not be compared across independently normalized
episodes as an absolute monetary or latency estimate.
\section{Pretraining Data and Training Details}
\label{app:training}

\subsection{Pretraining Data Sources}

RouteFM is pretrained using four data sources: LLMRouterBench,
RouterBench, RouterEval, and MixInstruct. After preprocessing, the complete
pretraining collection contains 212,470 query--task instances and
1,727,809 valid query--model observations.

Table~\ref{tab:pretrain_sources} summarizes the sampling distribution used
during episodic pretraining.

\begin{table}[h]
\centering
\caption{Pretraining data sources and episode sampling probabilities.}
\label{tab:pretrain_sources}
\small
\begin{tabular}{lrr}
\toprule
Source & Queries / Rows & Sampling Prob. \\
\midrule
LLMRouterBench & 25,790 & 45\% \\
RouterBench & 36,497 & 25\% \\
RouterEval & 40,183 & 25\% \\
MixInstruct & 110,000 & 5\% \\
\bottomrule
\end{tabular}
\end{table}

MMR-Bench is not used as a pretraining source.

\subsection{Episodic Pretraining Construction}

Each pretraining episode is sampled from a single source and task. An episode
contains an anonymous candidate set, behavioral observations for each
candidate, and a disjoint set of target queries.

Specifically,
\[
M\in[6,24],
\qquad
T\in[16,32],
\]
and the per-candidate behavioral observation budget is sampled from
\[
K
\in
\{8,16,32,64,128,256,512,1024\}.
\]
Context and target query IDs are always disjoint.

We use two behavioral-context layouts. In the \emph{aligned-dense} layout,
all candidates are observed on the same $K$ queries. In the
\emph{aligned-sparse} layout, observations are drawn from a larger shared
query grid and only a subset of candidate--query cells is revealed. The two
layouts are sampled with probabilities $4/7$ and $3/7$, respectively.

Target episodes are sampled from three complementary regimes:
\begin{itemize}
    \item \textbf{Natural} ($65\%$): preserves the native query and winner
    distribution of the task;
    \item \textbf{Opportunity} ($20\%$): emphasizes candidate pools for which
    different models are useful on different queries;
    \item \textbf{Boundary} ($15\%$): emphasizes targets with relatively small
    gaps between the strongest candidates.
\end{itemize}

Candidate order is randomly permuted in every episode. Episodes with similar
context scales are batched together to reduce unnecessary padding, while
ordinary masks handle variation in the number of candidates, context
observations, and targets.

\subsection{Pretraining Curriculum}

RouteFM is trained as a single continuous 10,000-step optimization trajectory
from random initialization. The architecture is fixed throughout training;
the curriculum changes only the context regime and the weight of the
routing-aware objective.

\begin{table}[h]
\centering
\caption{RouteFM pretraining curriculum.}
\label{tab:curriculum}
\small
\begin{tabular}{llll}
\toprule
Stage & Steps & Context Regime & Regret Weight \\
\midrule
I &
1--3,500 &
$K\in\{128,256,512,1024\}$ &
$0$ \\
II &
3,501--7,000 &
50\% large, 50\% $K\le64$ &
$0.15$ \\
III &
7,001--10,000 &
30\% large, 70\% $K\le64$ &
$0.25$ \\
\bottomrule
\end{tabular}
\end{table}

The first stage emphasizes stable capability estimation from relatively rich
behavioral evidence. The later stages progressively increase the frequency
of limited-context episodes and place greater weight on the routing decision
itself.

\subsection{Optimization and Loss Configuration}

We optimize RouteFM using AdamW with learning rate $10^{-4}$, weight decay
$0.01$, gradient clipping at $1.0$, and automatic mixed precision. The
microbatch size is $4$, with gradient accumulation yielding an effective
batch size of $16$.

Quality and relative cost are supervised using point-wise and pairwise terms.
The point-wise objective uses Smooth L1 loss with $\beta=0.05$,
\[
\mathcal{L}_{\mathrm{point}}
=
\ell_H(\hat y,y)
+
0.25\,
\ell_H(\hat c,\tilde c).
\]
Pairwise score and cost targets are constructed using temperature $0.05$,
while predicted pairwise differences use temperature $0.1$. The base
objective is
\[
\mathcal{L}_{\mathrm{base}}
=
\mathcal{L}_{\mathrm{point}}
+
0.5\mathcal{L}_{\mathrm{rank}}^{y}
+
0.1\mathcal{L}_{\mathrm{rank}}^{c}.
\]

For target query $t$, the routing distribution is
\[
\pi_{t,m}
=
\frac{
\exp(\hat y_{t,m}/0.1)
}{
\sum_{j\in\mathcal{M}_t}
\exp(\hat y_{t,j}/0.1)
},
\]
and the routing-aware regret objective is
\[
\mathcal{L}_{\mathrm{regret}}
=
\frac{1}{T}
\sum_t
\left[
\max_{m\in\mathcal{M}_t} y_{t,m}
-
\sum_{m\in\mathcal{M}_t}
\pi_{t,m}y_{t,m}
\right].
\]
The final objective is
\[
\mathcal{L}
=
\mathcal{L}_{\mathrm{base}}
+
\lambda_r\mathcal{L}_{\mathrm{regret}},
\]
where $\lambda_r$ is $0$, $0.15$, and $0.25$ in the three curriculum stages,
respectively. All loss terms are computed only over valid
target--candidate pairs.

\section{Evaluation Protocols}
\label{app:evaluation}

\subsection{Common Evaluation Rules}

Within each evaluation setting, all methods receive matched candidate pools,
behavioral-context queries, context budgets, target queries, and frozen query
representations. Target quality is revealed only after routing and is used
solely for evaluation.

The primary metric is routed quality, defined as the mean observed quality of
the model selected by the router. Unless explicitly stated otherwise, RouteFM
remains frozen and receives no target-domain parameter updates.

\subsection{In-Domain RouterEval Evaluation}

The in-domain evaluation is conducted on RouterEval tasks represented during
pretraining. We use three split seeds and construct four-candidate routing
pools for each task.

For each split, $40\%$ of queries are reserved as targets. Behavioral context
is sampled from the remaining queries using
\[
K\in\{8,16,32,64\}.
\]
Context and target query IDs are disjoint. All methods are evaluated using the
same candidate pools and query splits.

\subsection{MMR-Bench Cross-Modal Evaluation}

The multimodal evaluation uses the seven-dataset version of MMR-Bench:
MMStar, MathVerse, MathVision, MathVista, OCRBench, RealWorldQA, and
SEEDBench2-Plus. The benchmark contains 10,370 queries and nine complete
candidate models.

For RouteFM, the Qwen3-VL encoder jointly embeds the textual question and
associated image. Candidate answers, quality labels, and cost labels are never
included in the query embedding.

For the limited-context setting, each dataset is independently divided into five deterministic folds.
One fold serves as the target set, and behavioral context is sampled only from the remaining four
folds. The target fold is rotated over all five folds, so that every query appears exactly once as
a target during the five-fold evaluation. We evaluate
$K \in \{8, 16, 32, 64\}$,
with context sets nested across $K$.

For the large-context setting, each dataset is partitioned into $40\%$
behavioral context and $60\%$ target queries. Fixed-$K$ evaluation and the
40:60 protocol are treated as separate evaluation regimes.

\subsection{Target-Domain Support Protocol}

The pretraining ablation in Table~\ref{tab:mmrbench_ablation}(b) follows a
separate support-controlled protocol.

Each MMR-Bench dataset is first divided into $20\%$ labeled target-domain
support and an $80\%$ evaluation remainder. The evaluation remainder is then
split into a behavioral-context pool and a disjoint target set.

We compare:
\begin{enumerate}
    \item frozen pretrained RouteFM, without target-domain parameter updates;
    \item the same architecture trained from random initialization using only
    the labeled support set; and
    \item pretrained RouteFM fine-tuned using the same labeled support set.
\end{enumerate}

The two trainable variants use matched optimization budgets, episode draws,
candidate ordering, and validation rules. Neither behavioral-context labels
from the final evaluation split nor target labels are used for parameter
optimization or checkpoint selection.

Because this protocol differs from the standard MMR-Bench evaluation used in
Table~\ref{tab:mmrbench_ablation}(a), results should be compared within each
block rather than across the two blocks.

\subsection{New-Model Incorporation}

To evaluate dynamic candidate-pool expansion, each of the nine MMR-Bench
candidates is treated as the newly introduced model in turn. The eight
incumbent candidates receive $64$ behavioral observations, while the new
candidate receives
\[
K_{\mathrm{new}} \in \{0, 1, 2, 4, 8, 16, 32, 64\}.
\]

RouteFM remains frozen throughout the experiment. Trainable baselines are
refitted using the same behavioral evidence available in each setting.
Results are averaged over datasets, split seeds, and the nine choices of newly
introduced model.

The primary few-shot analysis focuses on $K_{\mathrm{new}} \geq 8$.
The $K_{\mathrm{new}} = 0$ and $K_{\mathrm{new}} = 1$ conditions are retained
as cold-start stress tests rather than evidence of reliable zero-shot model incorporation.

\subsection{Target-Domain Efficiency}

Target-domain efficiency is measured under the MMR-Bench large-context
protocol using precomputed query embeddings and behavioral observations.

For trainable baselines, reported latency includes both target-domain fitting
and batched target inference. RouteFM requires no target-domain fitting, so
only frozen batched inference is measured. Embedding extraction, candidate
answer generation, disk I/O, and one-time checkpoint loading are excluded.

Measurements are performed on a single NVIDIA A100 GPU using three warm-up
runs and 20 timed repetitions per dataset. The reported experiment therefore
measures adaptation-plus-routing latency after embeddings and behavioral
observations have already been obtained.

\subsection{Context-Budget Efficiency}

For each MMR-Bench dataset and evaluation split, the complete $40\%$
behavioral-context partition defines the $100\%$ context budget. RouteFM is
evaluated using nested prefixes from $1\%$ to $100\%$ while the target set is
held fixed.

For each competing router, we take its full-context routing quality as a
reference. We report the smallest tested context fraction at which RouteFM
reaches or exceeds this reference performance. This quantity measures the
minimum amount of behavioral context required for RouteFM to recover the
full-context routing quality of each competing method.

The shaded region in Figure~3(c) reports variation across the three independently
constructed splits used in this context-budget experiment.

\section{Baseline Implementations}

All baselines are adapted to the same anonymous, episode-local evaluation
protocol. They may use the behavioral-context embeddings and observed context
outcomes but never target outcomes.

\noindent \textbf{Context-Mean.}
Selects the candidate with the highest mean observed context quality and does
not use the target-query embedding.

\noindent \textbf{Weighted $k$NN.}
Uses cosine-similarity nearest neighbors with $k=10$ and softmax temperature
$0.07$ to estimate candidate quality from nearby behavioral observations.

\noindent \textbf{MLP.}
Uses a feed-forward predictor with hidden dimensions $256$ and $128$ and is
trained on the available behavioral context using mean-squared error.

\noindent \textbf{Embedding-Ridge.}
Fits an embedding-to-quality ridge predictor for each candidate. The
regularization coefficient is selected from
$\{0.01,0.1,1,10\}$ using context-only validation.

\noindent \textbf{EmbedLLM.}
Uses a learned query projection together with candidate representations and
their multiplicative interaction to estimate target-specific candidate
quality.

\noindent \textbf{RM-Softmax.}
Fits an episode-local linear predictor using a softmax-weighted
routing-regret surrogate.

\noindent \textbf{TRouter.}
Constructs a task-level semantic representation from the available context
and combines it with candidate performance profiles.

\noindent \textbf{InferenceDynamics.}
Constructs an episode-local capability representation from available
task/category structure and observed context outcomes.

\noindent \textbf{UniRoute.}
Builds cluster-level candidate error profiles and softly assigns each target
query to semantic clusters.

\textbf{ICL-Router.}
As an exception to the shared query-representation setting used by the other baselines,
ICL-Router is reproduced with its original method-specific representation pipeline.
It estimates, separately for each candidate model, the probability that it can answer
a target query given its historical successes and failures. We reproduce the authors' two-stage training on their released data: a learned projector first aligns frozen Qwen3-Embedding-8B query representations with Qwen2.5-7B-Instruct, after which the projector and router are fine-tuned on binary performance profiles. At evaluation, we provide each candidate's outcomes on the available Context queries and select the model with the highest predicted probability of \texttt{Yes}. On MMR-Bench, this baseline uses query text without images and receives no target-domain parameter updates.

All methods use fixed hyperparameter configurations within each evaluation protocol.
Candidate pools, behavioral context, and target queries are matched across methods.
Query representations are also matched except for ICL-Router, which retains its
method-specific pretrained representation pipeline as described above.

\section{Additional Results}
\label{app:additional}

\subsection{Effect of Pretraining Data Composition}
\label{app:data_ablation}

We further study how the composition of routing environments used during
pretraining affects the learned routing capability. Our default mixture samples
LLMRouterBench, RouterBench, RouterEval, and MixInstruct with probabilities
$45\%/25\%/25\%/5\%$, respectively. We consider two types of ablations.
First, we remove each source individually and redistribute its sampling
probability among the remaining sources. Second, we retain all four sources but
vary their sampling ratios: \emph{Uniform} uses
$25\%/25\%/25\%/25\%$, \emph{MixInstruct-heavy} uses
$10\%/20\%/20\%/50\%$, and \emph{Validation-matched} uses
$40\%/15\%/40\%/5\%$, following the same source order. All variants use the
same architecture, training procedure, and total pretraining budget of 160K
episodes; only the source composition is changed.

Table~\ref{tab:data_ablation} reports transfer performance on MMR-Bench,
which is not included in pretraining. The default heterogeneous mixture
performs best across all evaluated context sizes. Removing LLMRouterBench
causes the largest degradation, reducing quality by 4.09 points at $K=8$ and
2.09 points at $K=64$. Removing RouterBench also consistently hurts transfer,
while removing RouterEval or MixInstruct leads to smaller but systematic
decreases. These results indicate that no single source fully determines
RouteFM's transfer ability, but the diversity and balance of pretraining
routing environments substantially affect how well the learned routing
capability generalizes.

Changing the sampling ratios leads to a similar pattern. Uniform sampling
remains competitive at larger context sizes but is weaker in the low-context
regime, while heavily upweighting MixInstruct substantially degrades transfer.
Overall, these results suggest that heterogeneous routing pretraining benefits
not only from the amount of data, but also from the composition of routing
environments encountered during training.

\begin{table}[t]
\centering
\caption{
Effect of pretraining data composition on MMR-Bench.
All variants are trained with the same total budget of 160K episodes and
differ only in the sampling distribution over pretraining sources.
}
\label{tab:data_ablation}
\small
\begin{tabular}{lcccc}
\toprule
Pretraining mixture & $K=8$ & $K=16$ & $K=32$ & $K=64$ \\
\midrule
Default mixture
& \textbf{0.7323} & \textbf{0.7457} & \textbf{0.7487} & \textbf{0.7504} \\

w/o RouterEval
& 0.7292 & 0.7420 & 0.7443 & 0.7453 \\

w/o MixInstruct
& 0.7273 & 0.7387 & 0.7421 & 0.7430 \\

w/o LLMRouterBench
& 0.6914 & 0.7143 & 0.7191 & 0.7295 \\

w/o RouterBench
& 0.7156 & 0.7239 & 0.7291 & 0.7311 \\
\midrule
Uniform mixture
& 0.7208 & 0.7359 & 0.7460 & 0.7468 \\

MixInstruct-heavy
& 0.7168 & 0.7297 & 0.7407 & 0.7435 \\

Validation-matched
& 0.7282 & 0.7393 & 0.7466 & 0.7460 \\
\bottomrule
\end{tabular}
\end{table}

\subsection{Quality--Cost Trade-off}
\label{app:quality_cost}

\begin{figure}[htb]
    \centering
    \includegraphics[width=0.5\linewidth]{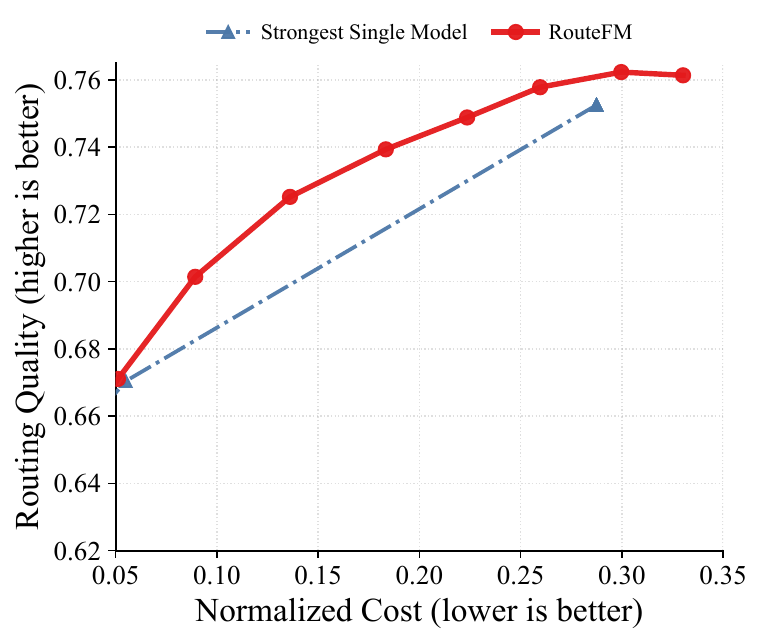}
    \caption{
    Quality--cost trade-off of RouteFM under different deployment preferences.
    Varying $\lambda$ in the routing objective
    $\hat y_m-\lambda\hat c_m$ produces different operating points without
    updating RouteFM's parameters. The single-model baseline shows the
    corresponding quality--cost operating points obtained without routing.
    }
    \label{fig:quality_cost}
\end{figure}

RouteFM predicts both target-specific quality and relative cost, allowing the
deployment preference $\lambda$ in the routing objective
$\hat y_m-\lambda\hat c_m$ to be adjusted without retraining the router.
Figure~\ref{fig:quality_cost} shows the resulting quality--cost trade-off.
By varying $\lambda$, a single frozen RouteFM spans a broad range of operating
points, from low-cost routing to higher-quality configurations. Across most of
this range, RouteFM achieves a more favorable quality--cost trade-off than
relying on a single model, demonstrating that the pretrained router can adapt
to different deployment preferences through the routing objective alone.

\section{Reproducibility and Development Details}

The RouteFM implementation and supporting materials are publicly available at
\url{https://github.com/LAMDA-Model-Reuse/RouteFM}.
The released model checkpoints are available at
\url{https://huggingface.co/AIGNLAI/RouteFM}.
The repository provides the model architecture, pretraining pipeline,
MMR-Bench evaluation scripts, fixed split definitions, and utilities for
evaluating custom anonymous candidate pools.

\section{Future Directions}
\label{app:future}

RouteFM suggests a broader view of LLM routing in which routing is treated as
a reusable capability rather than a policy specialized to a particular
deployment. This perspective naturally opens several directions for future
research.

First, future work may explore routing pretraining at a substantially broader
scale. Increasing the diversity of tasks, model families, modalities, and
deployment environments may enable routing models to acquire more general
capability representations and improve transfer to previously unseen settings.
As routing ecosystems continue to expand, it would also be interesting to
study how scaling properties emerge with respect to the diversity and coverage
of routing environments.

Second, reusable routing models may benefit from richer forms of deployment
context. The present formulation characterizes candidate models primarily
through their observed behavior, while practical routing decisions may depend
on additional signals such as latency, availability, resource constraints, or
other system-level preferences. A more general routing model could potentially
integrate such heterogeneous contextual information within a unified
decision-making interface.

Finally, an important long-term direction is to move from static routing
environments toward continuously evolving routing systems. In realistic
deployments, candidate models, query distributions, and operational conditions
may change over time. Developing routing models that can continually interpret
new evidence while preserving a reusable pretrained routing capability could
provide a foundation for more adaptive and long-lived model-routing
infrastructure.

\end{document}